\documentclass[10pt,twocolumn,letterpaper]{article}

 \usepackage{iccv}              

\definecolor{iccvblue}{rgb}{0.21,0.49,0.74}
\usepackage[pagebackref,breaklinks,colorlinks,allcolors=iccvblue]{hyperref}

\usepackage{tabularx}

\usepackage{array}  
\newcolumntype{C}[1]{>{\centering\arraybackslash}m{#1}}  
\usepackage{xcolor}
\usepackage{graphicx} 
\usepackage{array}    

\def\paperID{*****} 
\def\confName{ICCV}
\def\confYear{2025}

\title{Look Beyond Saliency: Low-Attention Guided Dual Encoding for Video Semantic Search}

\author{
Faisal Alhajari \and
Mohammed A. Alkhrashi \and
Alreem Almuhrij \and
Sarah Abuhimed \and
Noorh Aldossary \and
Abdullah Aldwyish \and
Raied Aljadaany \and
Huda Alamri \and
Muhammad Kamran J Khan\\
Saudi Data And Artificial Intelligence Authority (SDAIA)\\
{\tt\small fhajari, mkhrashi, amuhrij, T25-1885, T25-1884, aaldwyish, raljadaany, haamri, mkkhan@ncai.gov.sa}
}

\begin{document}
\maketitle

\begin{abstract}
Video semantic search in densely crowded scenes remains a challenging task due to visual encoders' tendency to prioritize salient foreground regions while neglecting contextually important, background areas. We propose an Inverse Attention Embedding mechanism that explicitly captures and highlights these overlooked regions. By combining inverse attention embeddings with traditional visual embeddings, our method significantly enhances semantic retrieval performance without additional training. Initial experiments and ablation studies demonstrate promising improvements over existing approaches in recall for video semantic search in crowded environments.
\end{abstract}
\vspace{-1em}    
\section{Introduction}
\label{sec:intro}

Video semantic search is a challenging multimodal task that enables users to retrieve relevant video frames based on natural language queries, such as \textit{a kid wearing pink} or \textit{a black bag left unattended}. This capability is particularly important in densely crowded environments, including public gatherings, transportation hubs, and major events, where accurate retrieval is crucial, yet difficult, due to occlusions, overlapping entities, and subtle visual details.

Current approaches typically encode both natural language queries and video frames into a common high-dimensional embedding space \cite{deng2023prompt,luo2021clip4clip,bain2021frozen}. During inference, user queries (e.g., \textit{a person riding a red bicycle in the rain}) and frame embeddings are compared in this shared semantic space, enabling retrieval of relevant frames even when exact keywords are absent. However, these methods often perform poorly in crowded scenes because visual encoders prioritize prominent, high-attention regions, neglecting contextually critical but subtle or occluded low-attention areas.

In this paper, we address the task of video semantic search specifically in densely crowded environments, such as public gatherings, transportation hubs, and large events. These environments pose unique challenges because visual encoders typically prioritize highly prominent visual regions, causing them to overlook regions with lower visual saliency or attention. However, these neglected low attention areas often contain essential semantic details such as partially occluded objects or subtle activities that are crucial for accurate video retrieval in crowded scenarios. Consequently, standard embeddings from visual encoders become incomplete or biased, negatively impacting retrieval performance.

To tackle this issue, we introduce a novel \textbf{Inverse Attention Region Embedding Cropping method}. Our approach explicitly targets and encodes these low-attention regions alongside standard visual embeddings, significantly enriching the semantic representation. By directly addressing the imbalance in attention, our technique effectively enhances the retrieval accuracy for crowded scenes without requiring additional model training.

\begin{figure}
    \centering
    \includegraphics[width=0.8\linewidth]{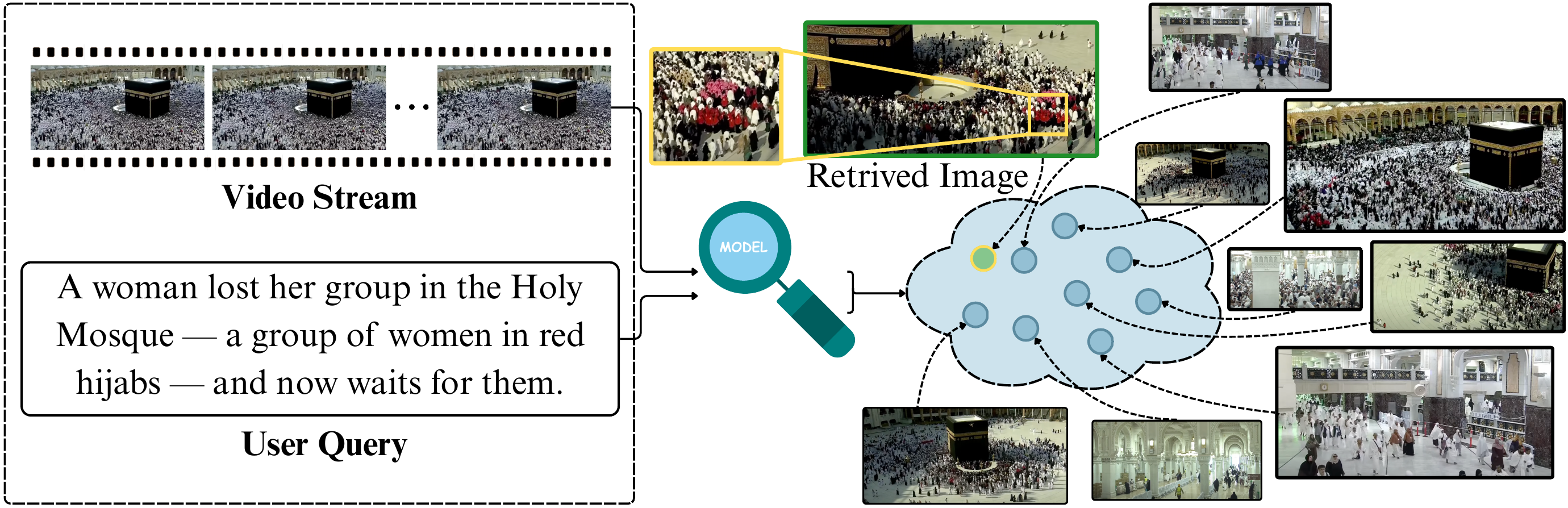}
    \caption{Given a natural-language query, for example: \textit{a group of women in red hijabs}, our model retrieves relevant content even in low-attention areas.
    }
    \label{fig:arch}
\end{figure}
The main contributions of this work are:
\begin{itemize}

\item Proposing \textbf{Inverse Attention Region Embedding Cropping}, a novel training-free method to embed overlooked low-attention regions.
\item Proposing \textbf{Dense-Set} a crowded subset of MS-COCO \cite{lin2014microsoft} that highlighting existing retrieval models failures in densely crowded scenes.
\item Evaluating existing retrieval models on densely crowded scenes.
\end{itemize}

\section{Related Work}

The video semantic search aims to retrieve relevant video content based on queries in natural language and has attracted great attention in the research of vision and language~\cite{radford2021learning,jia2021scaling,yao2021filip,gao2022pyramidclip,li2021align,luo2021clip4clip,bain2021frozen, ma2022xclip,gorti2022xpool}.
Many approaches build on dual-encoder architectures, where visual and textual inputs are projected into a shared embedding space using contrastive learning. Models such as CLIP\cite{radford2021learning} and ALIGN\cite{jia2021scaling} enable efficient inference-time retrieval by independently encoding each modality. Recent methods like FILIP\cite{yao2021filip} and PyramidCLIP\cite{gao2022pyramidclip} introduce finer-grained alignment by leveraging patch-level and hierarchical features. Cross-modal transformer models, such as ALBEF\cite{li2021align}, jointly encode image and text inputs through attention mechanisms, achieving stronger alignment at the cost of computational efficiency.

For video retrieval, models such as CLIP4Clip\cite{luo2021clip4clip}, Frozen in Time\cite{bain2021frozen}, and X-CLIP\cite{ma2022xclip} extend dual-encoder frameworks to temporal data by aggregating frame-level features and aligning them with sentence embeddings. Other approaches, like X-Pool \cite{gorti2022xpool}, introduce query-conditioned pooling to emphasize semantically relevant frames. While these methods perform well on standard datasets, they often prioritize high-salience visual content, which can lead to missed semantic signals,particularly in densely crowded environments, where subtle and low-attention regions may contain critical information often overlooked by existing approaches.

\section{Methodology}

\begin{figure*} [h!]
    \centering
    \includegraphics[width=0.8\linewidth]{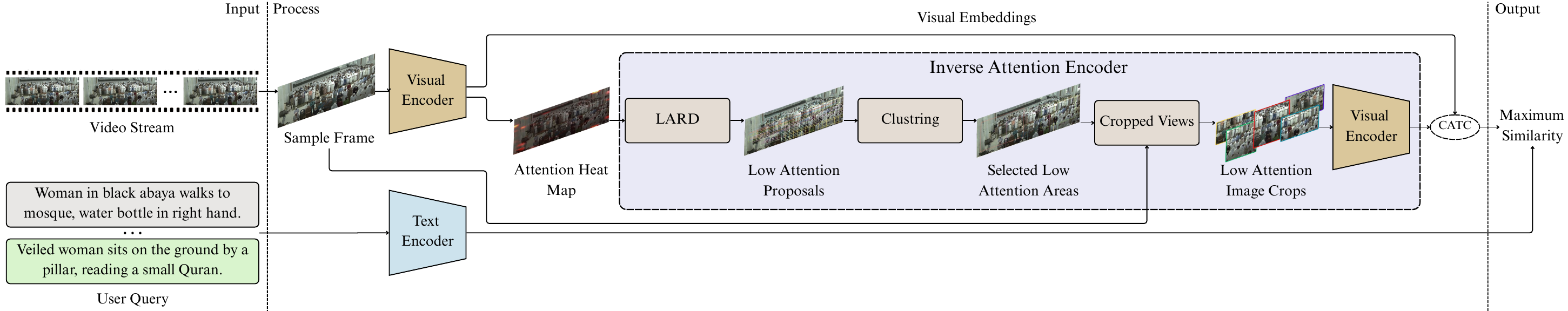}
    \caption{The overall workflow of our approach of Video Semantic Search via Inverse Attention Encoding, the key components and their interactions. LARD (Low-Attention Region Detector): yields a set of boxes that correspond to areas of the image that were underrepresented in the full image embedding. }
    \label{fig:arch}
\end{figure*}

Figure \ref{fig:arch} illustrates the workflow of our video semantic search method, the main component, we propose, in the pipeline is the Inverse-Attention Encoder module.
We adopt a dual-encoder architecture similar to CLIP~\cite{radford2021learning} and SigLIP~\cite{zhai2023sigmoid}, where a text encoder and a visual encoder independently project their respective inputs into a shared embedding space \(\mathbb{R}^D\). Given a video \(V = \{f_1, \dots, f_m\}\) consisting of \(m\) frames and a natural language query \(q = \{w_1, \dots, w_n\}\), the query is encoded as a global text embedding \(\mathbf{t} = A(q) \in \mathbb{R}^D\), and each frame \(f_i\) is encoded as a global visual embedding \(\mathbf{g}_i = V(f_i) \in \mathbb{R}^D\). While effective for large-scale retrieval, this approach tends to overlook subtle or occluded cues, especially in crowded scenes, due to its reliance on dominant attention signals.

To address this limitation, we introduce an \textit{Inverse Attention Encoder}, illustrated in Figure~\ref{fig:arch}, which augments standard frame-level embeddings with additional features derived from low-attention regions. For a given sampled frame \(f_r\), the visual encoder outputs both the global embedding \(\mathbf{g}\) and an attention heatmap \(h \in \mathbb{R}^{H \times W}\), highlighting the regions that most influenced the embedding.

We identify low-attention proposals using a component we call \textbf{LARD} (Low-Attention Region Detector), which thresholds the heatmap to generate a binary mask. A \(64 \times 64\) sliding kernel moves over this mask, and regions with mean attention below a fixed threshold \(T = 0.4\) are marked as candidate bounding boxes:
\[
Bb_i = \mathrm{LARD}(h).
\]

Next, these bounding boxes are grouped using \textbf{LARC} (Low-Attention Region Clustering), where k-means is used to form \(n\) larger and more coherent regions (we set \(n = 5\) in our implementation):
\[
\overline{Bb}_j = \mathrm{LARC}(Bb_i).
\]

Each of the resulting regions is passed to \textbf{LACE} (Low-Attention Crop Extractor), which crops and resizes the region from the original frame:
\[
C_j = \mathrm{LACE}(\overline{Bb}_j, f_r).
\]

Each crop \(C_j\) is re-encoded using the same visual encoder to obtain region-level embeddings \(\mathbf{r}_j = V(C_j) \in \mathbb{R}^D\). To integrate both local fine-grained and global contextual features, region-level embeddings are combined with the original global embedding using \textbf{LAFM} (Low-Attention Feature Merger):
\[
\overline{\mathbf{v}} = \mathrm{LAFM}(\mathbf{g}, \mathbf{r}_1, \dots, \mathbf{r}_n).
\]

To compute similarity with the text query, we calculate cosine similarity between the text embedding \(\mathbf{t}\) and each visual embedding, taking the maximum across all regions and the global embedding:
\[
S = \max \left\{ \cos(\mathbf{t}, \mathbf{g}), \max_j \cos(\mathbf{t}, \mathbf{r}_j) \right\}.
\]

This similarity score reflects the best alignment between the text and any part of the frame, global or localized. By incorporating low-attention content, our method recovers semantically meaningful cues typically overlooked in crowded environments. Crucially, all processing is done at inference time, without any additional training or fine-tuning of the underlying encoder.

\section{Experiments}
 \label{sec:experiments}

 We evaluated our proposed method on three datasets: MS-COCO, MS-COCO Dense-Set and Holy Mosque (Haram). Dense-Set and Holy Mosque are curated datasets designed for this work to better capture crowded and complex visual scenes. More details about them are provided in the following sections. We compare against standard CLIP-based models and a grid cropping baseline, using recall as the primary evaluation metric.

    
    \textbf{Holly Mosque Dataset:}
    A custom dataset collected from surveillance and scene footage of the Holy Mosque crowded scene. It consists of annotated image-text pairs designed to evaluate model performance in domain-specific retrieval and understanding tasks.
    To evaluate the performance of the methods, we compiled a dataset of 500 images captured by multiple cameras across the Holly Mosque in Makkah. 
    From the 500 images 169 (33.8\%) were annotated with an object name or an action in the image. Some images were associated with multiple captions, resulting in 277 image-text pairs. 
    
    \textbf{MS-COCO Dense-Set:}
    We create Dense-Set, a subset of MS-COCO aimed at benchmarking text-to-image retrieval methods in challenging settings. The dataset was created by selecting the top 10\% crowded images in terms of objects present (based on the 81 object classes of COCO). For each image, we choose the associated description to be one of the object classes in the image, specifically, an object class that is present exactly once in the image. 
    
    This makes the image-description pairs in Dense-Set very challenging, as each description can be contained in multiple images, and each description is not fully descriptive. We argue that this better mimics real-world text-to-image retrieval settings. And we show that existing methods underperform when benchmark against Dense-Set.

    We first evaluated the OpenAI CLIP model with VIT base patch size 16 on the Holy Mosque crowded scene data, followed by Jina CLIP v2 \cite{koukounas2024jinaclipv2multilingualmultimodalembeddings}. Then we tried SigLIP with different settings, standard, with Grid Crops, and finally our approach in which we concatenate inverse attention embeddings with standard visual encoder embeddings to focus on low attention areas.
    
\section{Results}
\label{sec:formatting}

\begin{table}[t!]
\centering
\caption{Semantic Search Performance: Recall scores on MS-COCO, the Holy Mosque and MS-COCO Clutter datasets}
\label{tab:Tab2}
\footnotesize
\begin{tabular}{@{}l@{\hspace{0.4em}}cc@{}@{\hspace{0.4em}}cc@{}@{\hspace{0.4em}}cc@{}}
\toprule
\textbf{Model}  & \multicolumn{2}{c}{\textbf{MS-COCO}} & \multicolumn{2}{c}{\textbf{Holy Mosque}} & 
\multicolumn{2}{c}{\textbf{Dense-Set}}
\\
 \cmidrule(lr){2-3} \cmidrule(lr){4-5} \cmidrule(lr){6-7}
                                &\textbf{R@5} & \textbf{R@10}
                                &\textbf{R@5} & \textbf{R@10}
                                &\textbf{R@5} & \textbf{R@10}  
                                \\
\midrule
OpenAI CLIP         &62.5&72.8 &     31.0     &      31.0 
& 38.9 & 39.2      
\\
Jina CLIP v2 &65.3&75.5&     25.0     &      23.0
& 41.5& 40.9   
\\

SigLIP   &\textbf{74.6}&\textbf{82.6}& 34.0 & 36.0
        &40.6&41.0
\\
SigLIP + Grid Crops          &
67.4 & 76.7 
& 34.0          & 34.0  
& 39.8 & 43.11       
\\
ours &66.3&74.7& \textbf{41.0} & \textbf{47.0}  
&  \textbf{42.5}    &     \textbf{44.3}      
\\

\bottomrule
\end{tabular}
\end{table}

\begin{table*}
  \centering
  \footnotesize
  \setlength{\tabcolsep}{3pt}
\begin{tabular}{
  C{0.02\linewidth}  
  C{0.326\linewidth}  
  C{0.326\linewidth}  
  C{0.326\linewidth}  
}
\toprule
\textbf{Stage} & \textbf{Holy Mosque Scene 1} & \textbf{Holy Mosque Scene 2} &  \textbf{MS-COCO}\\
\midrule
\rotatebox{90}{\textit{Input frame}} &
\includegraphics[width=0.95\linewidth]{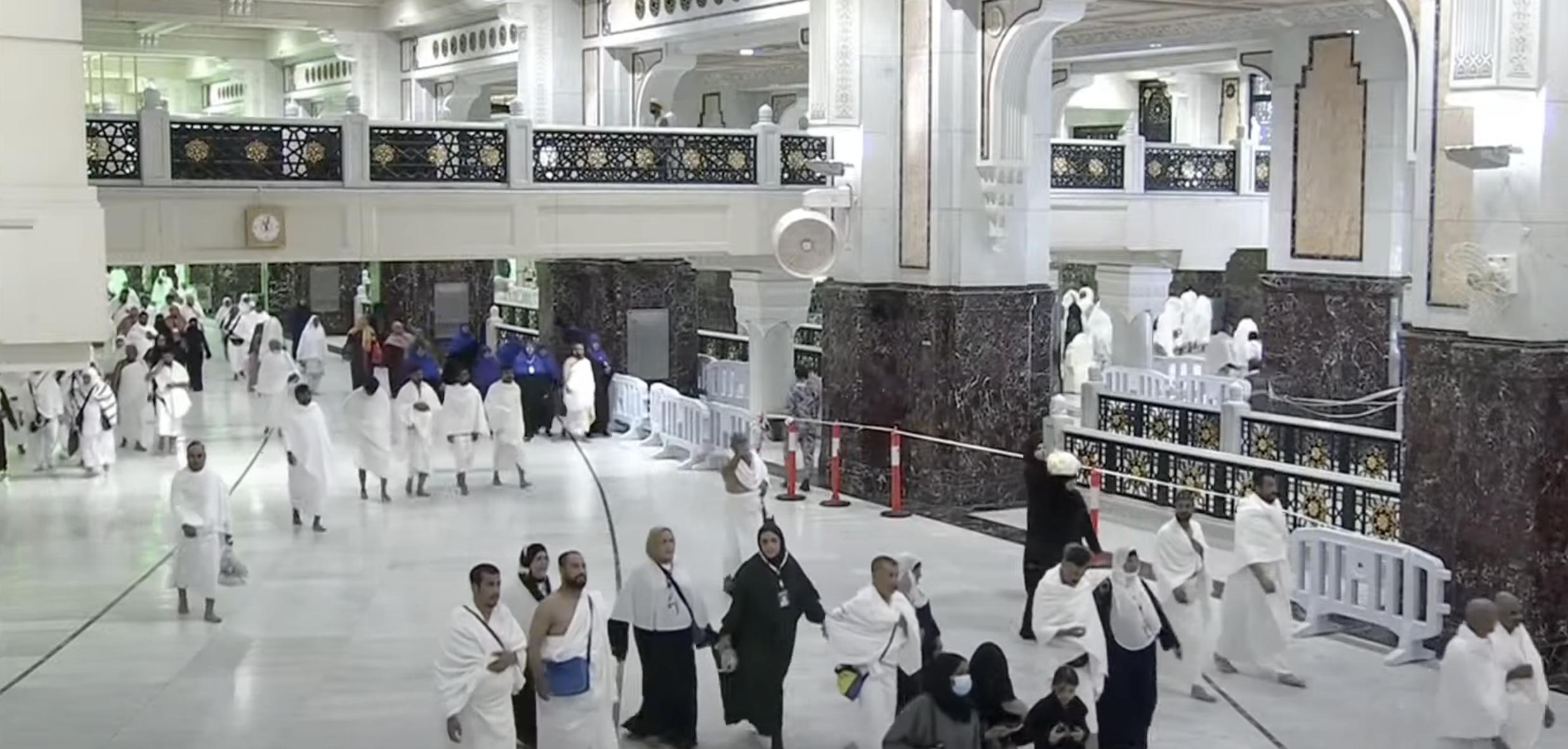} &
\includegraphics[width=0.99\linewidth]{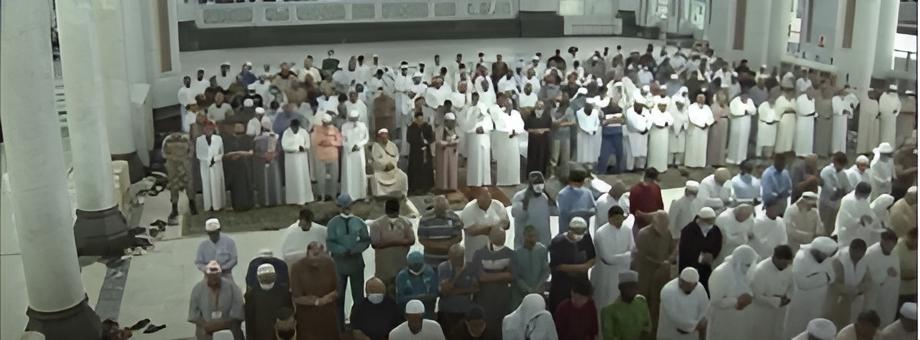} &
\includegraphics[width=0.70\linewidth]{figures/T3/ms_coco_example/original.jpg} \\


\rotatebox{90}{\textit{LARD}} &
\includegraphics[width=0.95\linewidth]{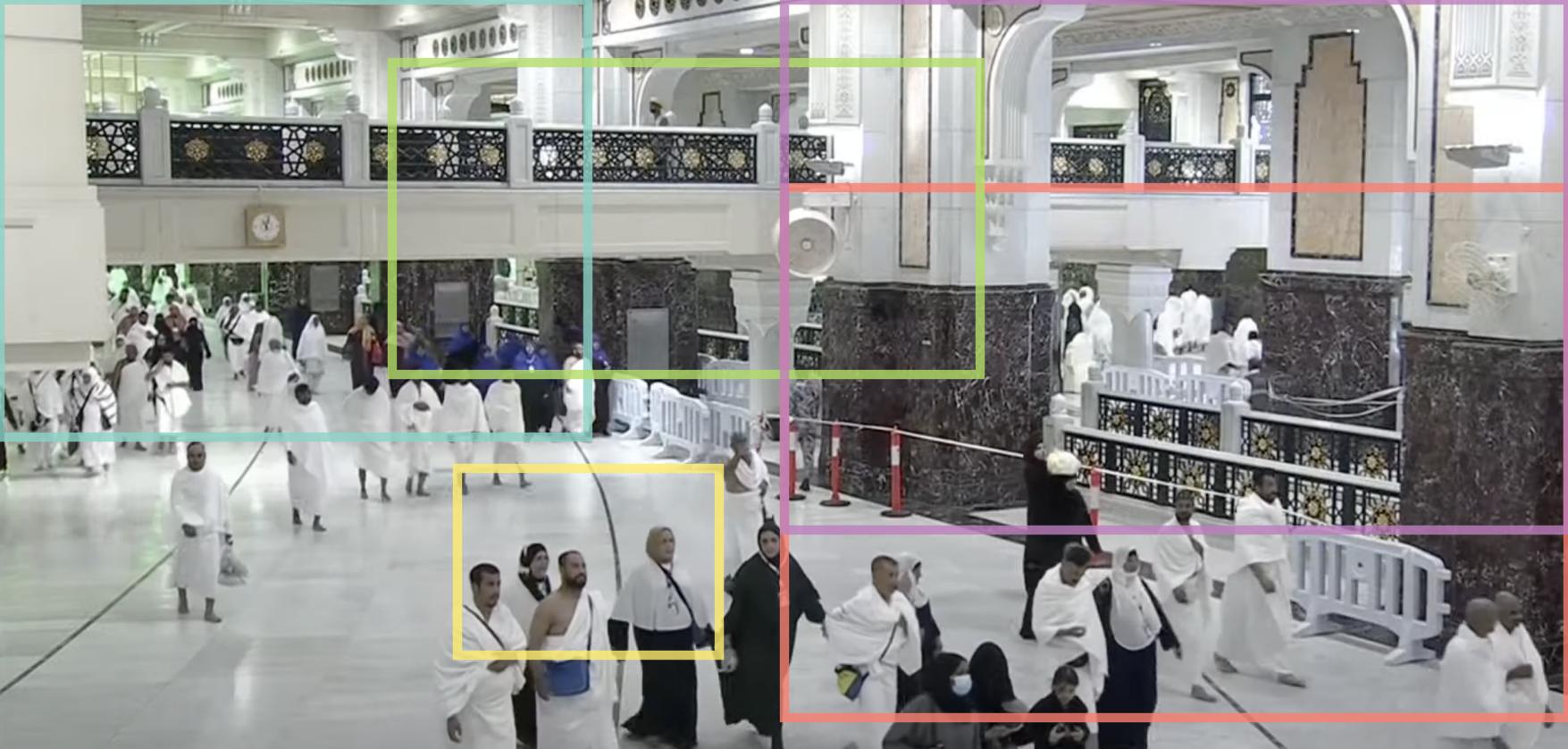} &
\includegraphics[width=0.99\linewidth]{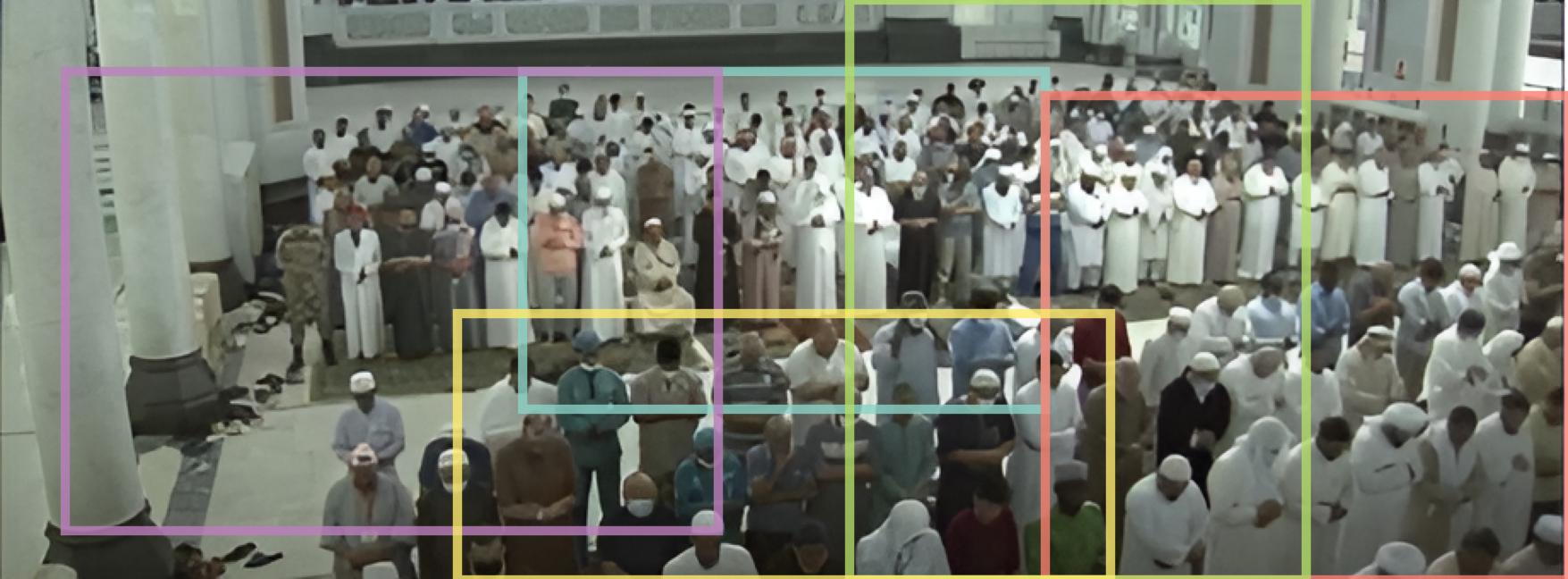} &
\includegraphics[width=0.70\linewidth]{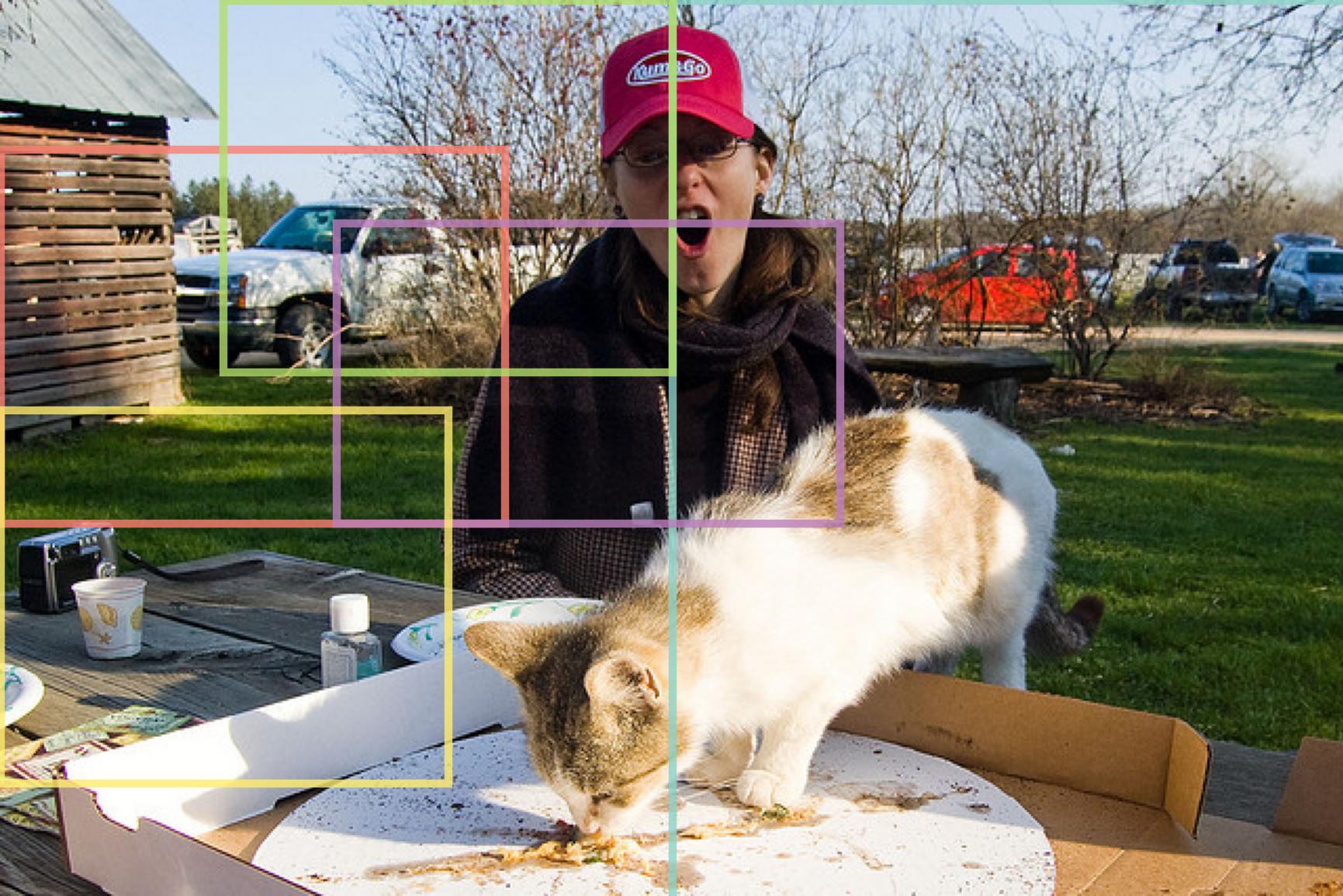} \\

\rotatebox{90}{\textit{Image Crops}} &
\includegraphics[width=0.95\linewidth]{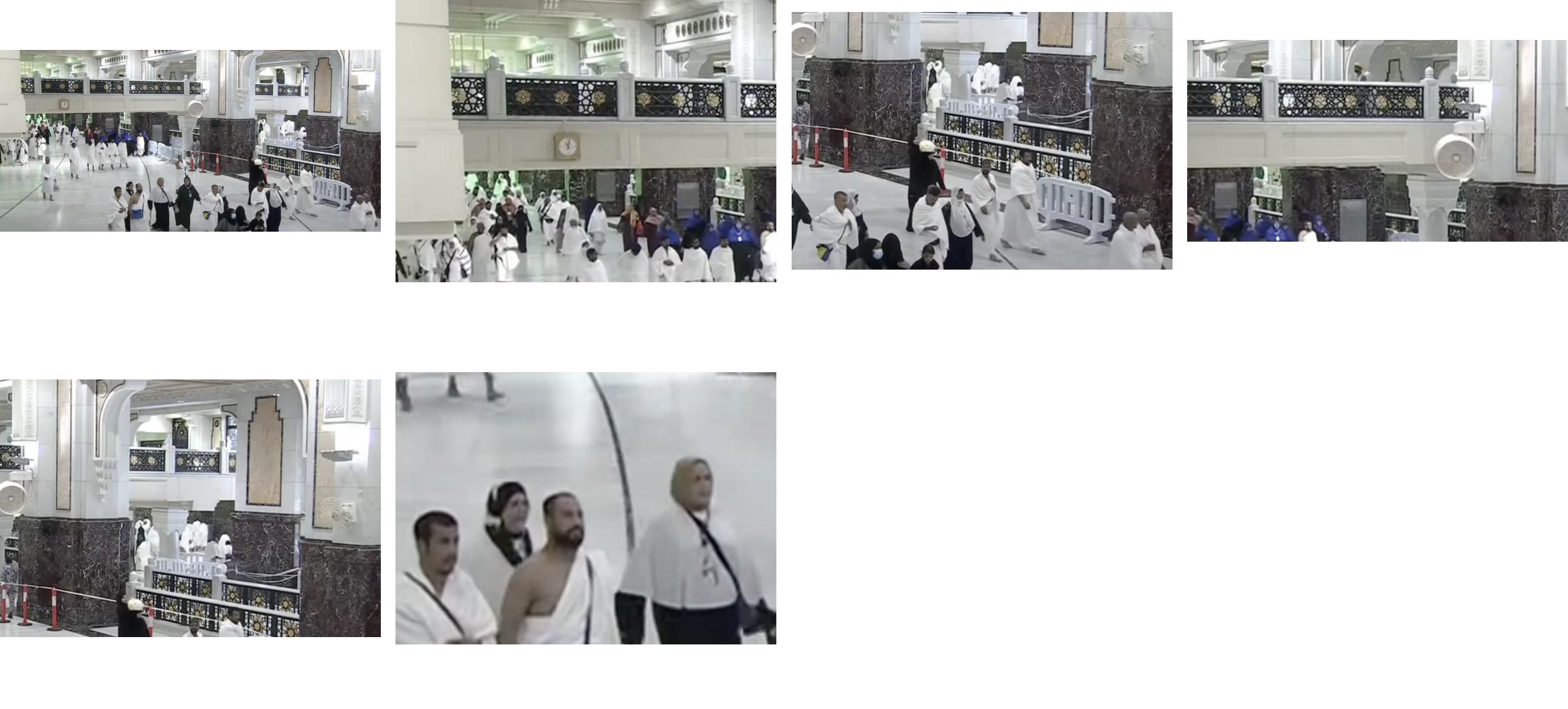}  &
\includegraphics[width=0.99\linewidth]{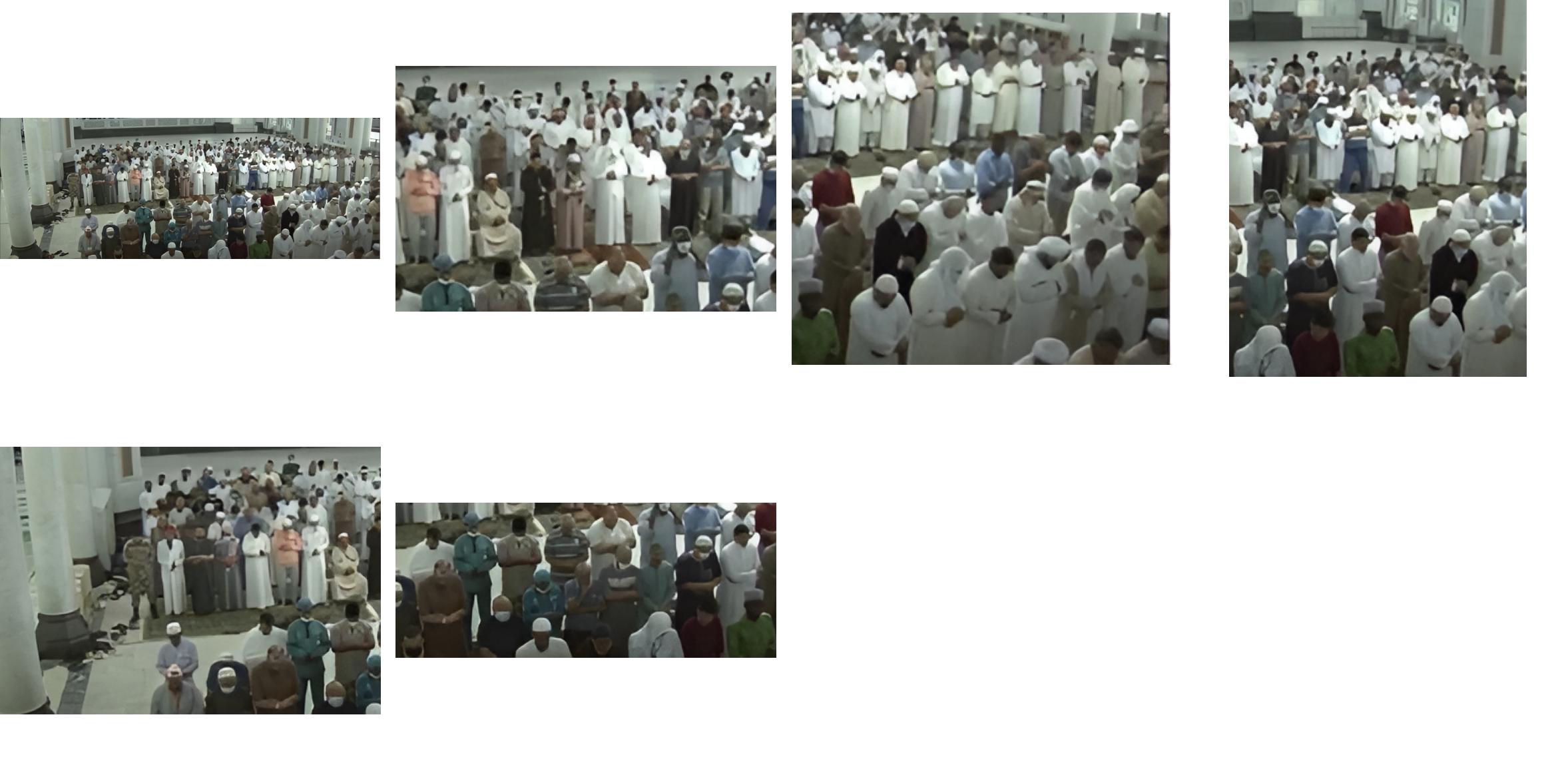}  &
\includegraphics[width=0.70\linewidth]{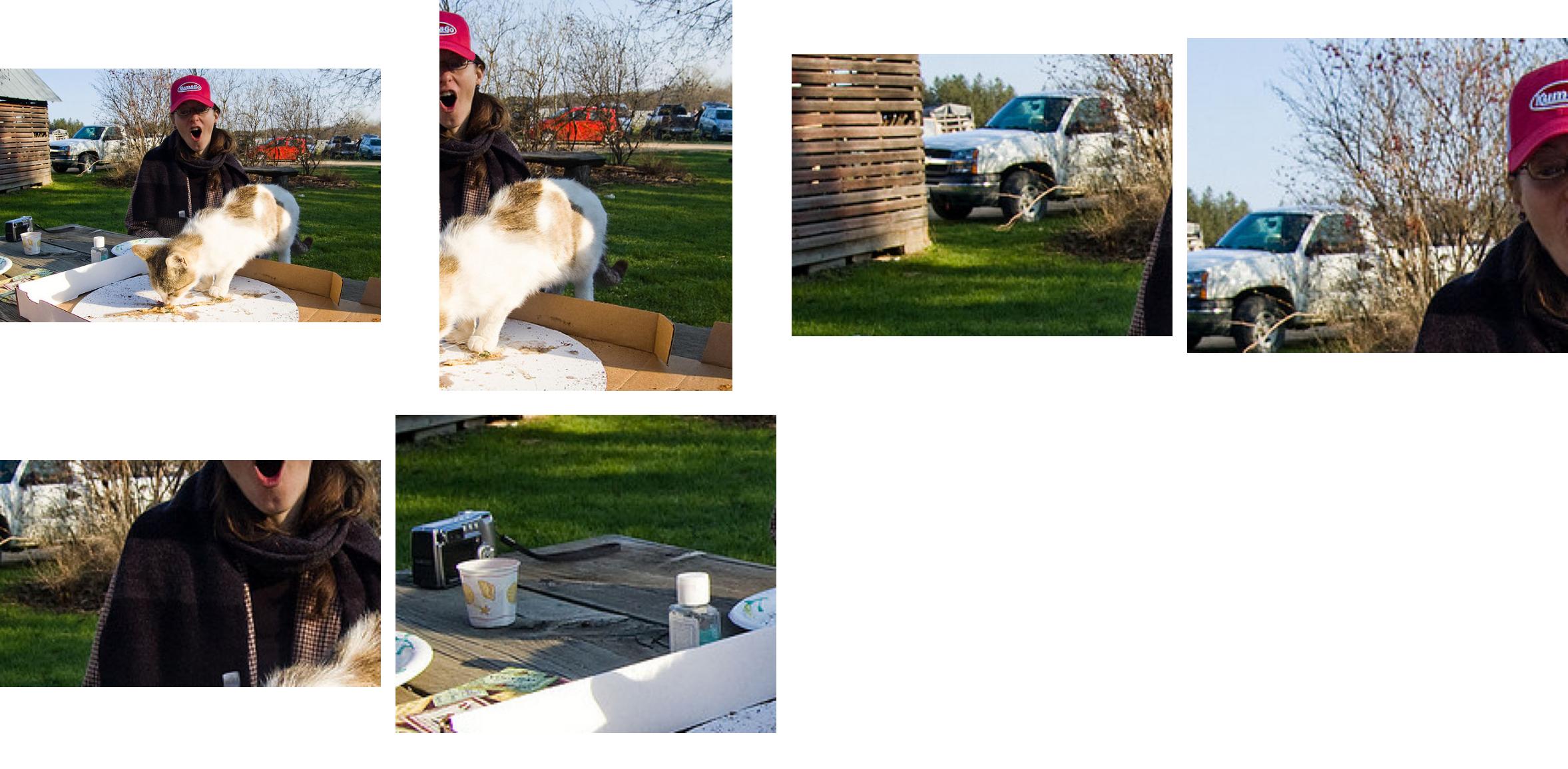}  \\

\rotatebox{90}{\textit{Retrieval Similarity}} &
\includegraphics[width=0.95\linewidth]{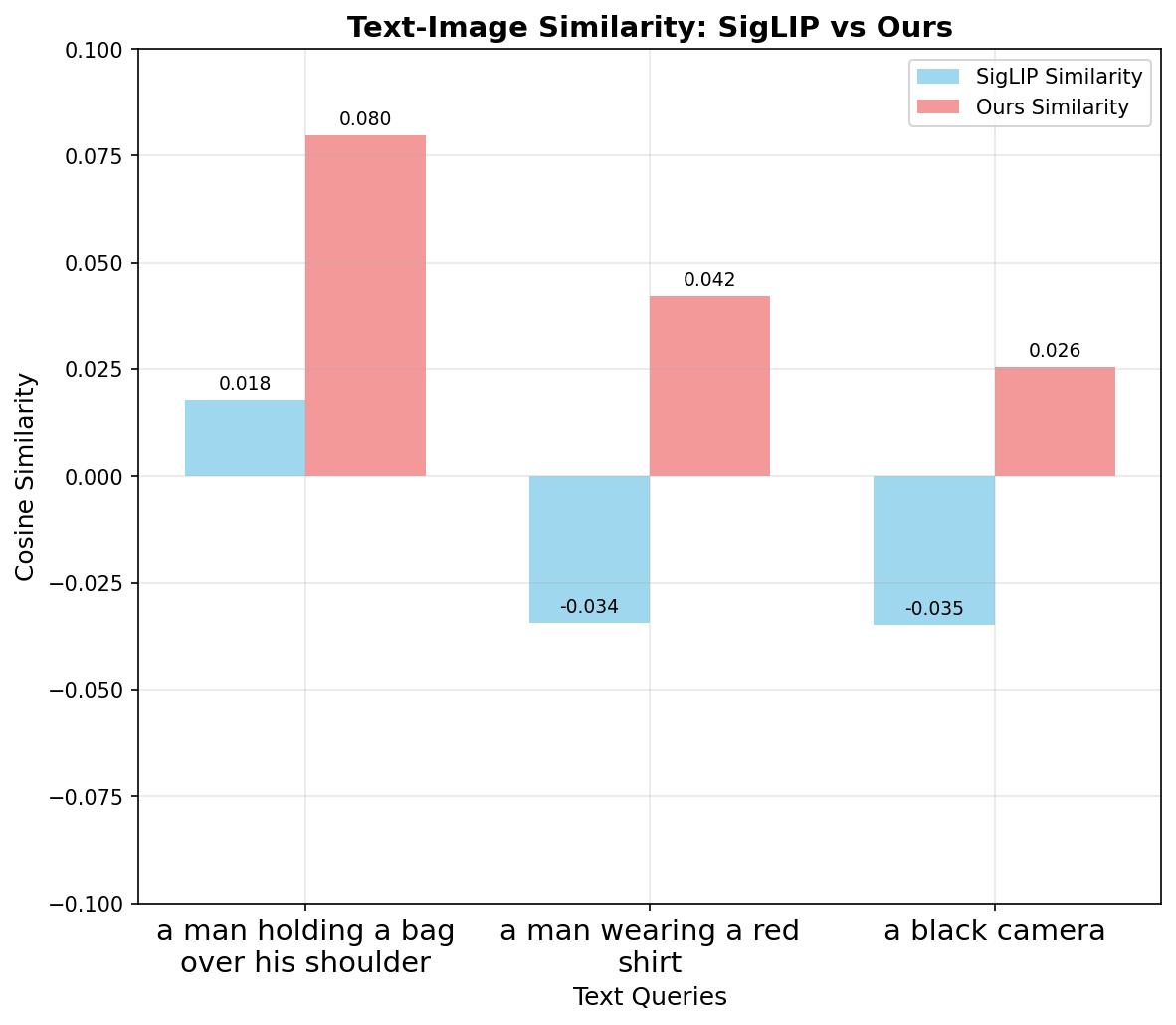} &
\includegraphics[width=0.99\linewidth]{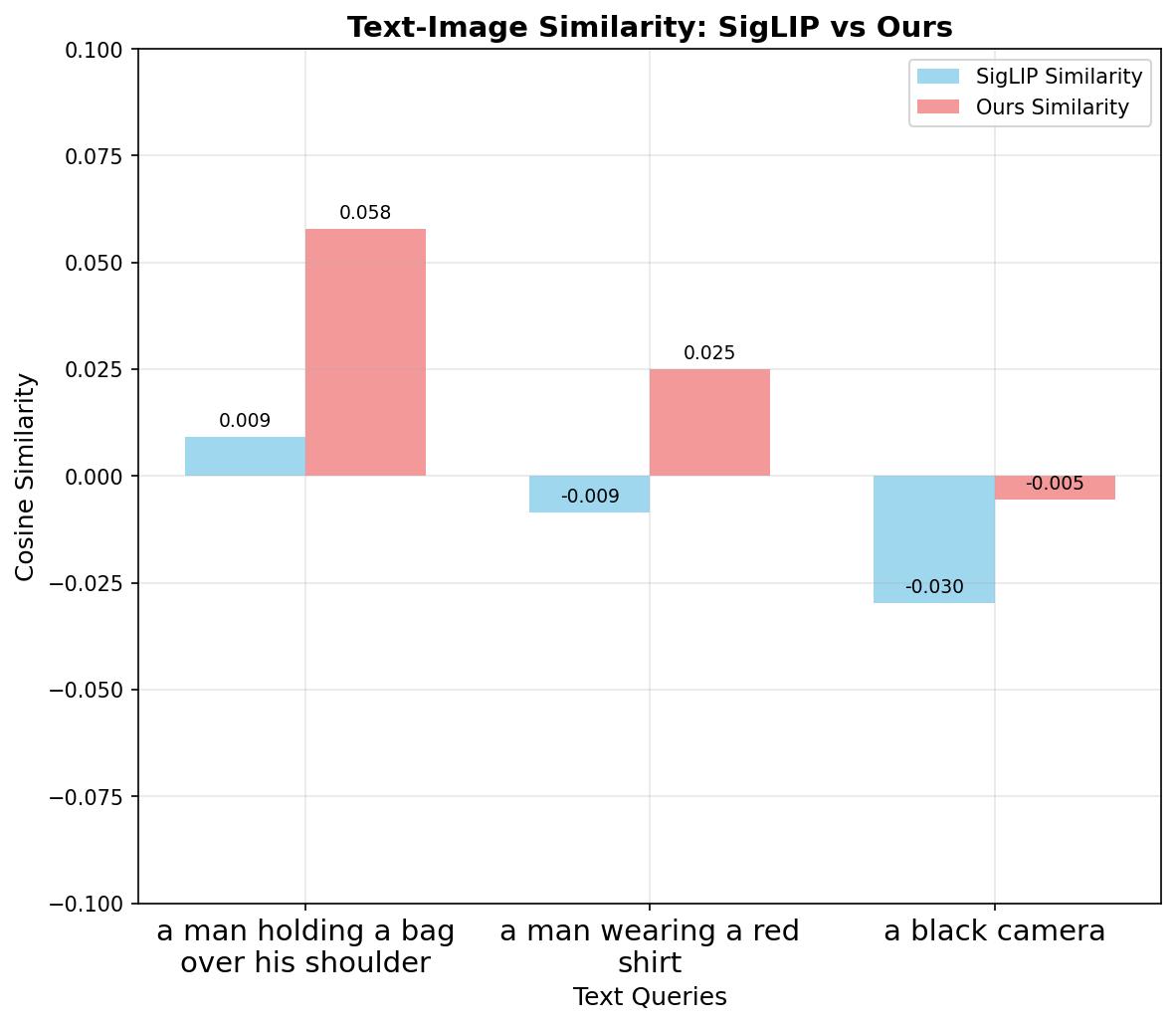} &
\includegraphics[width=0.95\linewidth]{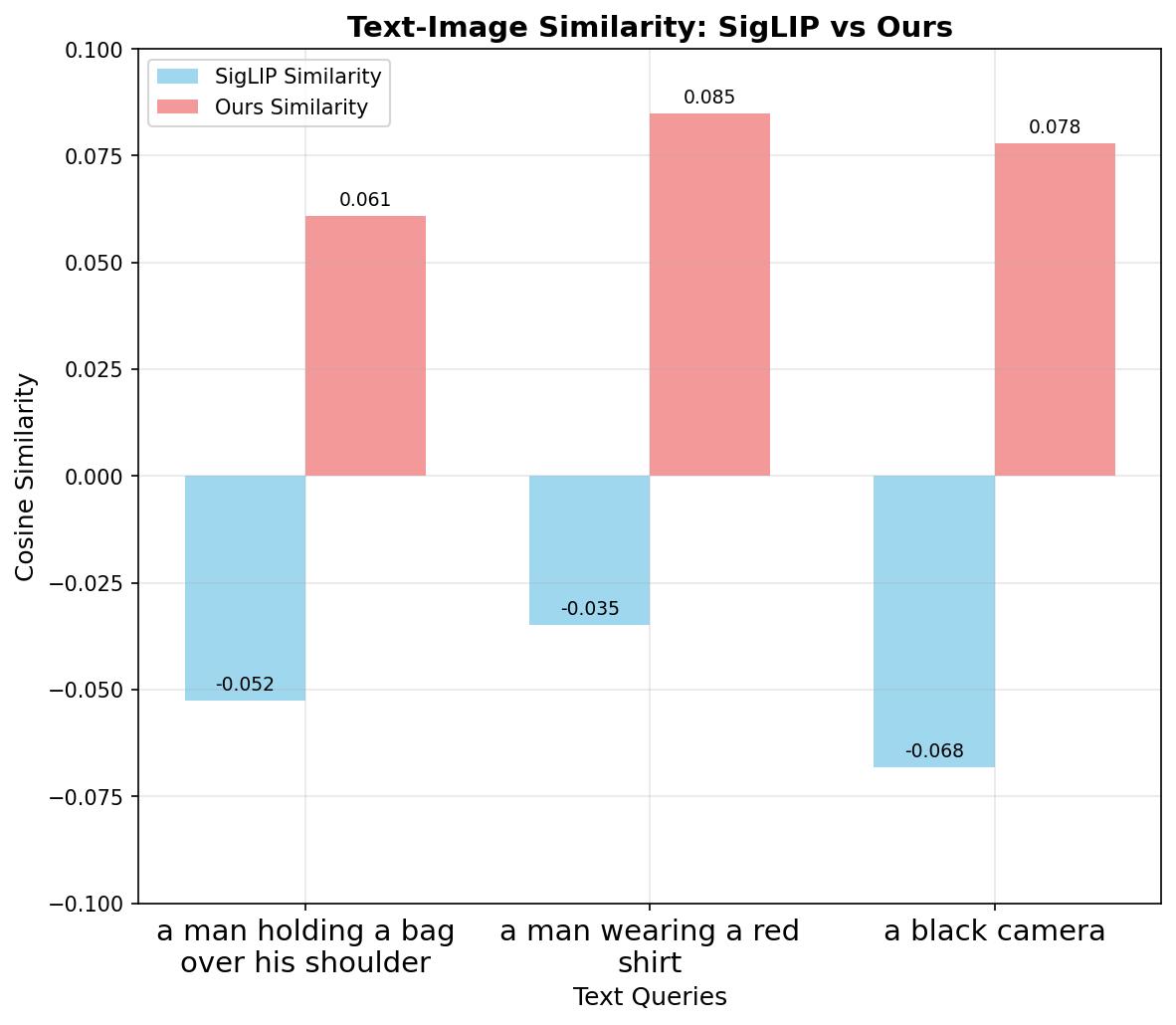}  \\
\bottomrule
\end{tabular}

\caption{Qualitative results of Video Semantic Search (our model) on different crowd scenes and viewpoints. From top to bottom: Input frame, Low Attention Regions, Image Cropped views, and the retrieval similarity scores of our model vs SigLIP.}
\label{tab:pipeline_results}
\end{table*}

Table \ref{tab:Tab2} presents the semantic search performance measured by recall scores in the MS-COCO dataset, the Holy Mosque, and Dense-Set datasets.



\textbf{Standard Encoding:}
The openAI CLIP \cite{ilharco_gabriel_2021_5143773} version of the SigLIP model \cite{zhai2023sigmoid} is evaluated in the Holy Mosque dataset without any fine-tuning, as this model showed the best zero-shot performance on the Holy Mosque dataset. This is used as a baseline for the ability to improve the zero-shot capability of CLIP based models on crowded scenes without training.

\textbf{Encoding Grid Crops:}
A naive approach is to crop each image into equal sized crops. then embeds each crop with addition to the full image and take the maximum similarity between the crops and the text query to increase the the performance on small objects. Grid crops does exactly that. and is used to showcase the importance of finding a more sophisticate approach to find better more meaningful crops. 


\textbf{Inverse Attention Encoding:}
The experimental results highlight the limitations of standard CLIP and SigLIP models  \cite{zhai2023sigmoid}  in crowded scenarios. Our proposed pipeline outperforms the baseline on crowded scenes like in the Holy Mosque dataset and Dense-Set. 
Table~\ref{tab:Tab2} demonstrates the recall scores for video semantic search in three benchmark datasets. MS-COCO, Dense-Set and the Holy Mosque dataset. The effectiveness of the inverse attention embedding mechanism is witnessed by achieving the highest recall on the Crowded Scene dataset. A R@5 of 0.41 and a R@10 of 0.44 on the Holy Mosque dataset and a R@5 of 0.42 and R@10  0.44 on the Dense-Set datasets, representing significant performance improvements over existing methods to handle complex and densely populated scenarios.
The ground truth captions in the MS-COCO dataset usually represents the entire image, not focusing on the single object, because of this, our model slightly under-performs other methods.
In open source datasets, our model is comparable to existing models, indicating reasonable generalization performance on standard benchmarks. 
These results validate the design of our pipeline as a more robust way to enhance the performance of semantic search models without training them on crowded scenes.


\textbf{Visual Results:}
The table \ref{tab:pipeline_results} showcases visual results for the retrieval task on the Holy Mosque and MS-COCO datasets. Each column presents an example from a dataset, and each row represents a stage in the retrieval process. The top row shows the input frame; the second highlights low-attention regions ignored during embedding. The third row displays cropped views of these regions, revealing small but relevant objects. The bottom row compares the cosine similarity of our model with SigLIP, where our model drastically increases the similarity for the queries about fine-grain background objects e.g., a man wearing a bag. 

\section{Conclusion }

We introduce an Inverse Attention Encoding mechanism to improve video semantic search in crowded scenes by incorporating low-attention regions typically overlooked by visual encoders. Our approach operates entirely at inference time and requires no additional training. Initial experiments on challenging datasets demonstrate improved retrieval performance. In future work, we plan to explore fine-tuning the visual encoder jointly with the proposed inverse attention mechanism and enhancing the text encoder by grounding textual representations in the semantics of the extracted low-attention regions.
{
    \small
    \bibliographystyle{ieeenat_fullname}
    \bibliography{main}
}

\end{document}